\documentclass[10pt, a4paper]{article}
\usepackage{lrec}
\usepackage{multibib}
\newcites{languageresource}{Language Resources}
\usepackage{graphicx}
\usepackage{tabularx}
\usepackage{soul}
\usepackage{graphicx,bm,booktabs}
\usepackage{multirow}
\usepackage{amsmath,paralist}
\usepackage{epstopdf}
\usepackage[latin1]{inputenc}

\usepackage{hyperref}
\usepackage{xstring}

\title{Towards Neural Speaker Modeling in Multi-Party Conversation:\\ The Task, Dataset, and Models}

\name{Zhao Meng,$^{1,2}$\quad Lili Mou,$^{3}$\quad Zhi Jin$^{1,*}\!\!\!$ \thanks{$^*$Corresponding author.}}

\address{$^1$Key Laboratory of High Confidence Software Technologies, MoE; Software Institute, Peking University\\
	$^2$Department of Computer Science, ETH Zurich\\
	$^3$  AdeptMind Research, Toronto, Canada \\
	zhmeng@student.ethz.ch, doublepower.mou@gmail.com,  lili@adeptmind.ai, zhijin@sei.pku.edu.cn}

\abstract{Neural network-based dialog systems are attracting increasing attention in both academia and 
	industry. Recently, researchers have begun to realize the importance of speaker modeling in neural dialog systems, but there lacks established tasks and datasets.
	In this paper, we propose \textit{speaker classification} as a surrogate task for general speaker modeling, and collect massive data to facilitate research in this direction. We further investigate temporal-based and content-based models of speakers, and propose several hybrids of them. Experiments show that speaker classification is feasible, and that hybrid models outperform each single component. \\ \newline  }

\begin{document}
\maketitleabstract

	\section{Introduction}
	
	
	Human-computer conversation has long attracted attention in both academia and industry.
	Researchers have developed a variety of approaches, ranging from rule-based systems for task-oriented
	dialog \cite{task_dialog_1,task_dialog_2} to data-driven models for open-domain conversation~\cite{SMTdialog}.
	
	A simple setting in the research of dialog systems is context-free, i.e., only a single utterance is considered during reply generation~\cite{NRM}. Other studies leverage context information by concatenating several utterances~\cite{context_dialog} or building hierarchical models~\cite{hierarchical_1,hierarchical_2}. The above approaches do not distinguish different speakers, and thus speaker information would be lost during conversation modeling.
	
	Speaker modeling is in fact important to dialog systems, and has been studied in traditional dialog research. However, existing methods are usually based on hand-crafted statistics and \textit{ad hoc} to a certain application~\cite{walker}. Another research direction is speaker modeling in a multi-modal setting, e.g., acoustic and visual~\cite{multiparty}, which is beyond the focus 
	of this paper.
	
	Recently, neural networks have become a prevailing technique in both task-oriented and open-domain dialog systems. After single-turn and multi-turn dialog studies, a few researchers have realized the role of speakers in neural conversational models. \newcite{persona} show that, with speaker identity information, a sequence-to-sequence neural dialog system tends to generate more coherent replies.
	In their approach, a speaker is modeled by a learned vector (also known as an \textit{embedding}). Such method, unfortunately, requires massive conversational data for a particular speaker to train his/her embedding, and thus does not generalize to rare or unseen speakers.
	
	\newcite{addressee} formalize a new task of addressee selection on online forums: by leveraging either the temporal or utterance information, they predict whom a post is talking to. While tempting for benchmarking speaker modeling, the task requires explicit speaker ID mentions, which occurs occasionally,  and thus is restricted.
	
	In this paper, we propose a \textit{speaker classification} task that predicts the speaker of an utterance. It serves as a surrogate task for general speaker modeling, similar to \textit{next utterance classification} \cite[NUC]{ubuntu} being a surrogate task for dialog generation. 
	The speaker classification task could also be useful in applications like \textit{speech diarization},\footnote{Speech diarization aims at answering ``who spoke when''~\cite{diarization}.}
	where text understanding can improve speaker segmentation, identification, etc.~in speech processing~\cite{textimproves,textimproves2}.
	
	We further propose a neural model that combines temporal and content information with interpolating or gating mechanisms. The observation is that, what a speaker has said (called {\tt content}) provides non-trivial background information of the speaker. Meanwhile, the relative order of a speaker (e.g., the $i$-th latest speaker) is a strong bias: nearer speakers are more likely to speak again; we call it {\tt temporal} information.
	We investigate different strategies for combination, ranging from linear interpolation to complicated gating mechanisms inspired by Differentiable Neural Computers~\cite[DNC]{DNC}.

	To evaluate our models, we constructed a massive corpus using transcripts of TV talk shows from the Cable News Network website. 
	Experiments show that combining content and temporal information significantly outperforms either of them, and that simple interpolation is surprisingly more efficient and effective than gating mechanisms.
	
	Datasets and code are available on our project website.\footnote{\url{https://sites.google.com/site/neuralspeaker/}}

	\section{Task Formulation and Data Collection}

	We formulate speaker classification as follows. 
	
	Assume that we have segmented a multi-party conversation into several parts by speakers; each segment comprises one or a few consecutive sentences $u_1, u_2, \cdots, u_N$, uttered by a particular speaker. 
	A candidate set of speakers $\mathcal{S}=\{s_1, s_2,\cdots, s_k\}$ is also given. In our experiments, we assume $u_1, u_2, \cdots, u_N$'s speaker $s_i$ is in $\mathcal{S}$. The task of speaker classification is to identify the speaker $s_i$ of $u_1,\cdots, u_N$. 
	
	Following the spirit of distributed semantics (e.g., word embeddings), we represent the current utterance(s) as a real-valued vector $\bm u$ with recurrent neural networks. Speakers are also represented as vectors $\bm s_i, \cdots, \bm s_k$. The speaker classification is accomplished by a softmax-like function
	\begin{align}
	\widetilde p_i &= \exp\left\{\bm s_i^\top\bm u\right\}\\
	p(s_i)  &= \frac{ \widetilde p_i}{\sum_j \widetilde p_j} 
	\label{eqn:softmax}
	\end{align}
	Because the number of candidate speakers may vary, the ``weights'' of softmax are not a fixed-size matrix, but the distributed representations of candidate speakers, $\bm s_1,\cdots,\bm s_k$.
	In Section~\ref{s:approach}, we investigate several approaches of modeling $\bm s_i$ based on what a speaker says or the relative order of a speaker in the dialog; we also propose to combine them by interpolating or gating mechanisms.

	To facilitate the speaker classification task, we crawled transcripts of more than 8,000 episodes of TV talk shows.\footnote{\url{https://transcripts.cnn.com}} We assumed that the current speaker is within the nearest $k$ speakers. ($k=5$, but at the beginning, $k$ may be less than 5.) Since too few utterances do not provide much information, we required each speaker having at least 3 previous utterances, but kept at most 5. Samples failing to meet the above requirements were filtered out during data preprocessing. 
	
	We split train/val/test sets according to TV show episodes instead of sentences; therefore no utterance overlaps between training and testing. Table~\ref{tab:stat} shows the statistics of our dataset partition. 
	
	\begin{table}\centering
		\resizebox{.7\linewidth}{!}{\begin{tabular}{lr}
				\toprule
				\textbf{Data partition} & \textbf{\# of samples} \\
				\midrule
				Train      &174,487 \\
				Validation & 21,071 \\
				Test       &  20,501 \\
				\bottomrule
			\end{tabular}
		}
		\caption{Dataset statistics.}\label{tab:stat}
	\end{table}
	
	\begin{figure*}[!t]
		\centering
		\includegraphics[width=.71\textwidth]{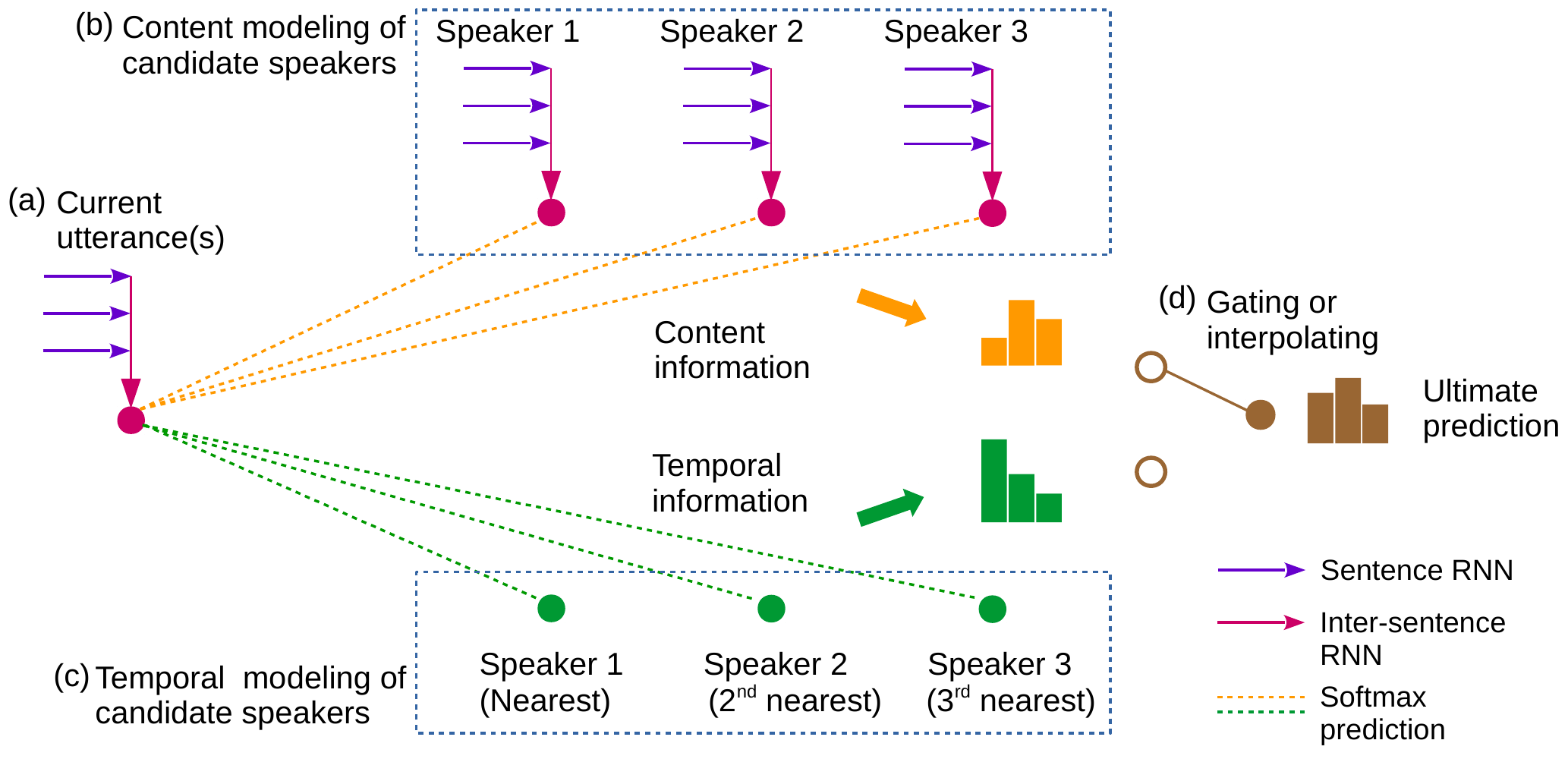}
		\caption{Hybrid content- and temporal-based speaker classification with a gating mechanism.}
		\label{fig:model}
	\end{figure*}

	\section{Methodology}\label{s:approach}
	
	We use a hierarchical recurrent neural network~\cite{hierarchical_2} to model the current utterances 
	$u_1,\cdots, u_N$ (Figure~\ref{fig:model}a). In other words, a recurrent neural network (RNN) captures the meaning of a sentence; another LSTM-RNN aggregates the sentence information into a fixed-size vector.  For simplicity, we use RNN's last state as the current utterances' representation ($\bm u$ in Equation~\ref{eqn:softmax}).
	
	In the rest of this section, we investigate content-based and temporal-based prediction in Subsections~\ref{ss:content} and~\ref{ss:temporal}; the spirit is similar to ``dynamic'' and ``static'' models, respectively, in \newcite{addressee}. We combine content-based and temporal-based prediction using gating mechanisms in Subsection~\ref{ss:gating}.
	
	\subsection{Prediction with Content Information}\label{ss:content}

	In this method, we model a speaker by what he or she has said, i.e., content.

	Figure~\ref{fig:model}b illustrates the content-based model: a hierarchical RNN (which is the same as Figure~\ref{fig:model}a) yields a vector $\bm s_i$ for each speaker, based on his or her nearest several utterances. The speaker vector $\bm s_i$ is multiplied by current utterances' vector $\bm u $ for softmax-like prediction (Equation~\ref{eqn:softmax}). We pick the candidate speaker that has the highest probability. 
	
	It is natural to model a speaker by his/her utterances, which provide illuminating information of the speaker's background, stance, etc. As will be shown in Section~\ref{sec:exp}, content-based prediction achieves significantly better performance than random guess. This also verifies that \textit{speaker classification} is feasible, being a meaningful surrogate task for speaker modeling.
	
	\subsection{Prediction with Temporal Information}\label{ss:temporal}
	\begin{table*}[!t]
		\centering
		\resizebox{.9\linewidth}{!}{
			\begin{tabular}{|ll||c|c|c||c|c|}
				\hline
				\multicolumn{2}{|l||}{\textbf{Model}} & \makebox[2.1cm]{\textbf{Macro} $F_1$}& \makebox[2.1cm]{\textbf{Weighted} $F_1$} & \makebox[2.1cm]{\textbf{Micro} $F_1$} & \makebox[2.1cm]{\textbf{Acc.}} & \makebox[2.1cm]{\textbf{MRR.}}  \\
				\hline
				\hline
				\multicolumn{2}{|l||}{Random guess} & 19.93 & 34.19 & 27.53 & 27.53 &  N/A \\
				\multicolumn{2}{|l||}{Majority guess} &21.26 & 62.96 & 74.01  & 74.01  & N/A \\
				\multicolumn{2}{|l||}{Hybrid random/majority guess} & 25.26 & 61.99 & 69.29  & 69.29 & N/A\\
				\hline
				\hline
				\multicolumn{2}{|l||}{Temporal information} & 26.07 & 63.60 & 73.99 & 73.99 & 84.85 \\
				\multicolumn{2}{|l||}{Content information} & 42.61 & 65.04 & 61.82 & 58.58 & 74.86 \\
				\multicolumn{2}{|l||}{\quad+ static attention} & 42.50 & 65.28 & 61.79 & 58.99 & 74.89 \\
				\multicolumn{2}{|l||}{\quad+ sentence-by-sentence attention} & 42.56 & 65.96 & 62.86 & 59.81 & 75.58 \\
				\hline
				\hline
				\multirow{3}{*}{\rotatebox{90}{\!\!Hybrid}}\!\!\!\! &Interpolating after training & \textbf{44.25}  & \textbf{71.35} &\textbf{76.10} & \textbf{75.84}& \textbf{85.73} \\
				&Interpolating while training &  41.30& 70.10&75.57 & 75.31 & 85.20 \\
				& Self-adaptive gating & 39.45 & 69.55 & 74.11 &74.09 & 84.85\\
				\hline
			\end{tabular}
		}
		\caption{Model performance (in percentage). }
		\label{tab:performance}
	\end{table*}
	In temporal-based approach, we sort all speakers in a descending order according to the last time he or she speaks, and assign a vector (embedding) for each index in the list, following the ``static model'' in \newcite{addressee}. Each speaker vector is randomly initialized and optimized as parameters during training. The predicted probability of a speaker is also computed by Equation~\ref{eqn:softmax}. 
	
	The temporal vector is also known as a \textit{position embedding} in other NLP literature~\cite{position}. 
	Our experiments show that temporal information provides strong bias: nearer speakers tend to speak more; hence, it is also useful for speaker modeling.
	
	\subsection{Combining Content and Temporal Information}\label{ss:gating}
	
	As both content and temporal provide important evidence for speaker classification, we propose to combine them by interpolating or gating mechanisms (illustrated in Figure~\ref{fig:model}d). In particular, we have
	\begin{equation}
	\bm p^\text{(hybrid)} = (1- g)\cdot \bm p^\text{(temporal)} +  g \cdot \bm p^\text{(content)}
	\label{eqn:predict}
	\end{equation}
	
	Here, $g$ is known as a \textit{gate}, balancing these two aspects. We investigate three strategies to compute the gate.
	
	\begin{enumerate}
		\item \textbf{Interpolating after training.} The simplest approach, perhaps, is to train two predictors separately, and interpolate after training by validating the hyperparameter $g$.
		\item \textbf{Interpolating while training.} We could also train the hybrid model as a whole with cross-entropy loss directly applied to Equation~\ref{eqn:predict}.
		\item \textbf{Self-adaptive gating.} 
		Inspired by hybrid content- and location-based addressing in Differentiable Neural Computers~\cite[DNCs]{DNC}, we design a learnable gate in hopes of dynamically balancing temporal  and content information. Different from DNCs, however, the gate here is not based on input (i.e., $\bm u$ in our scenario), but the result of content prediction $\bm p^\text{(content)}$. Formally
		\begin{equation}
		g = \operatorname{sigmoid} \left( w \cdot  \operatorname{std}[\, p^\text{(content)}\,]  + b \right) 
		\end{equation}
		where we compute the standard deviation ($\operatorname{std}$) of $ p$. 
		$w$ and $b$ are parameters to scale $\operatorname{std}[\, p^\text{(content)}\,]$ to a sensitive region of the sigmoid function.
	\end{enumerate}
	
	\section{Experimental Results}\label{sec:exp}

	\indent\textbf{Setup}. All our neural layers including word embeddings were set to 100-dimensional. We tried larger dimensions, resulting in slight but insignificant improvement. We did not use pretrained word embeddings but instead randomly initialized them because our dataset is large. We used the Adam optimizer~\cite{adam} mostly with default hyperparameters. We set the batch size to 10 due to GPU memory constraints. Dropout rate and early stop were also applied by validation. Notice that validation was accomplished by each metric itself because different metrics emphasize different aspects of model performance.
	
	\noindent\textbf{Performance}. Table~\ref{tab:performance} compares the performance of different models. 
	Majority-class guess results in high accuracy, showing that the dataset is screwed. Hence, we choose macro $F_1$  as our major metric, which addresses minority classes more than other metrics. We nevertheless present other metrics including accuracy, mean reciprocal ranking (MRR), and micro/weighted $F_1$ as additional evidence.
	
	As shown, the content-based model achieves higher performance in macro $F_1$ than majority guess, showing the effectiveness of content information. Following~\newcite{reasoning}, we adopt a static or sentence-by-sentence attention mechanism. The LSTM-RNN attends to speaker $ s_i$ to obtain speaker vector $\bm s_i$ while it is encoding current utterances. However, such attention mechanisms bring little improvements (if any). Hence, we do not use attention in our hybrid models for simplicity. 

	\begin{figure}[!t]
		\centering
		\includegraphics[width=.73\linewidth]{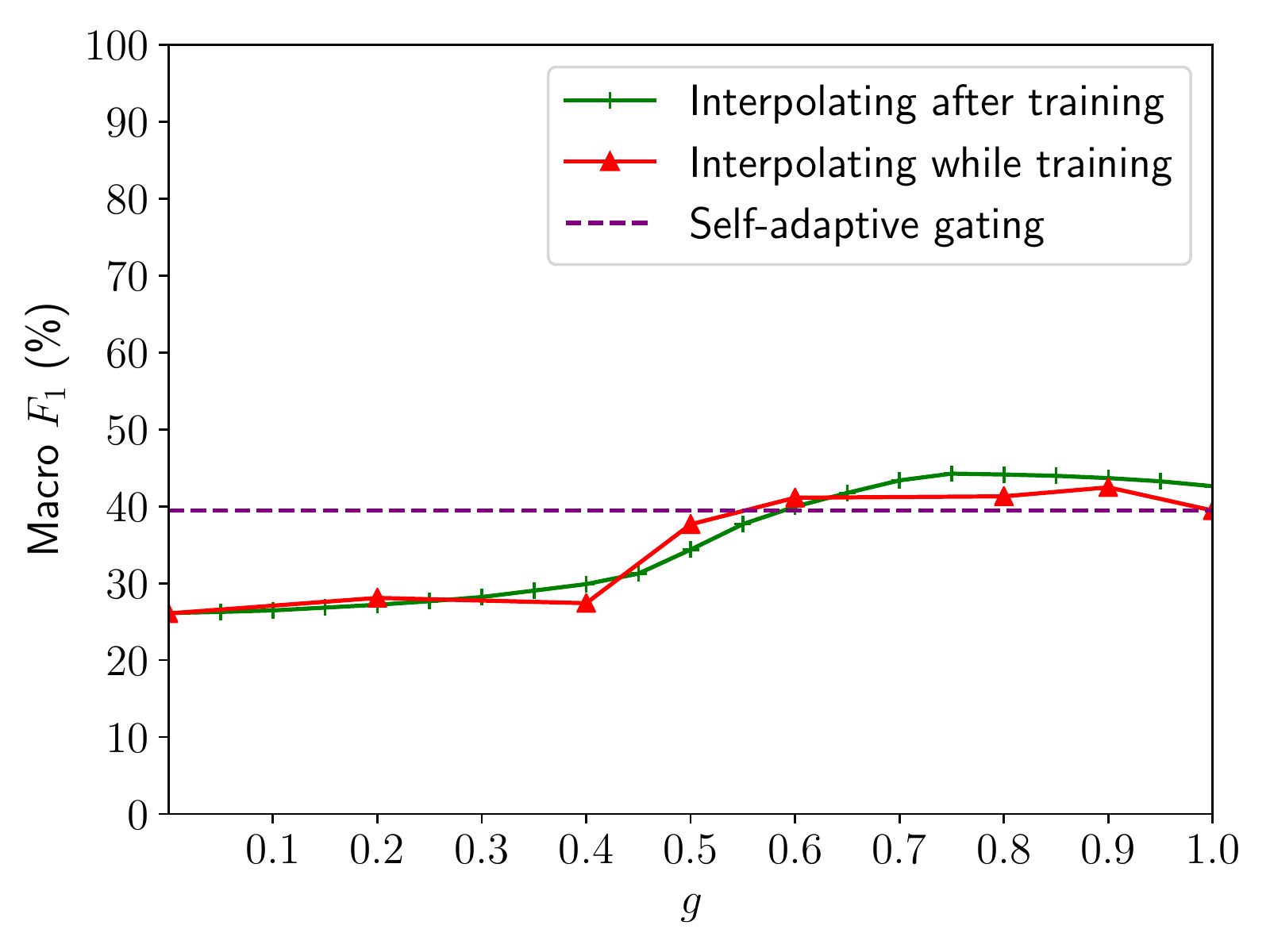}
		\caption{Performance vs.~the hyperparameter $g$ (solid lines). $g=0$: temporal only; $g=1$: content only. The self-adaptive gating mechanism is also plotted for comparison (which is not associated with a particular value of $g$).}
		\label{fig:analysis}
	\end{figure}
	
	All hybrid models achieve higher performance compared with either content- or temporal-based prediction in terms of most measures, which implies content and temporal information sources capture different aspects of speakers.

	Among different strategies of hybrid models, the simple approach ``interpolating after training'' surprisingly outperforms the other two. A plausible explanation is that training a hybrid model as a whole leads to optimization difficulty in our scenario; that simply interpolating well-trained models is efficient yet effective. However, the hyperparameter $g$ is sensitive and only yields high performance in the range $(0.6, 0.9)$. Thus, the learnable gating mechanism could also be useful in some scenarios, as it is self-adaptive.

	\section{Conclusion and Future Work}\label{s:conclusion}
	
	In this paper, we addressed the problem of neural speaker modeling in multi-party conversation. 
	We proposed \textit{speaker classification} as a surrogate task and collected massive TV talk shows as our corpus. We investigated content-based and temporal-based models, as well as their hybrids.
	Experimental results show that speaker classification is feasible, being a meaningful task for speaker modeling; that interpolation between content- and temporal-based prediction yields the highest performance.
	
	In the future, we would like to design more dedicated gating mechanisms to improve the performance; we would also like to explore other aspects of speaker modeling, e.g., incorporating dialog context before current utterances. The collected dataset is also potentially useful in other applications.

\section{Acknowledgments}
This research is supported by the National Basic Research Program of China (the 973 Program) under Grant
No.~2015CB352201, and the National Natural Science
Foundation of China under Grant Nos.~61232015 and
61620106007. An extended abstract of this paper appears as \newcite{AAAI} at AAAI-18 (to appear).

\section{Bibliographical References}
\label{main:ref}

\bibliographystyle{lrec}
\bibliography{xample}

\end{document}